%% file: main.tex
\documentclass[cameraready]{Interspeech}
\usepackage{multirow}
\usepackage{xcolor}
\usepackage[capitalise,noabbrev]{cleveref}

\title{Entity Binding Failures in Speech LLM Reasoning: Diagnosis and
Chain-of-Thought Intervention}

\author[affiliation={1}]{Ming-Hao}{Hsu}
\author[affiliation={2}]{Xiaohai}{Tian}
\author[affiliation={2}]{Jun}{Zhang}
\author[affiliation={1}, correspondingauthor]{Zhizheng}{Wu}
\address{
    $^1$ School of Data Science, The Chinese University of Hong Kong, Shenzhen, China \\
    $^2$ ByteDance, China
}

\email{hsuminghao1006@gmail.com, xiaohai171@gmail.com, zhangjun.jarry@bytedance.com, wuzhizheng@cuhk.edu.cn}

\keywords{speech language models, reasoning, entity binding, chain-of-thought, modality gap}

\usepackage{comment}

\begin{document}

\maketitle

\input{text/0_abstract}
\input{text/1_introduction}
\input{text/2_method}
\input{text/3_experiments}
\input{text/4_analysis}
\input{text/5_conclusion}
\section{Disclosure of AI Tool Use}
A generative AI assistant was used for editing and polishing the manuscript.
All experimental design, implementation, analysis, and scientific claims are the work of the authors.

\bibliographystyle{IEEEtran}
\bibliography{refs}

\end{document}

%% file: text/0_abstract.tex
\begin{abstract}
Speech Large Language Models (SLLMs) underperform their text counterparts on complex reasoning. We reveal that this gap is not a uniform cognitive deficit. Evaluating two architecturally diverse SLLMs, we show speech-to-text (S2T) matches or exceeds text-to-text (T2T) on spatial, syntactic, and factual tasks. Yet on logical tasks requiring entity tracking, S2T accuracy collapses to chance. We diagnose this as an entity binding failure: continuous speech features blur precise entity-property associations during implicit reasoning. To validate this diagnosis, we introduce Entity-Aware Chain-of-Thought (EA-CoT), a lightweight inference-time intervention forcing SLLMs to enumerate entities and bind them to claims before reasoning. EA-CoT bridges the gap, even when spoken names are misrecognized, yielding up to a 24.4 percentage-point accuracy gain. Ablations confirm the gains stem from explicit semantic binding, reframing the gap as an elicitation failure rather than a missing capability.
\end{abstract}

%% file: text/1_introduction.tex
\section{Introduction}
\label{sec:introduction}

Speech Large Language Models (SLLMs) can directly process spoken input, enabling natural conversational interactions without relying on cascaded text transcription \cite{rubenstein2023audiopalm, tang2024salmonn, zhang2023speechgpt, chu2023qwen}. However, recent benchmarking indicates that these models consistently underperform their text-based counterparts on complex reasoning tasks \cite{chen2024voicebench, yang2025sakura, lin2025vera, yang2024audiobench}. Previous studies have provided valuable macroscopic insights into this speech-to-text (S2T) versus text-to-text (T2T) modality gap, often attributing it to generalized information dilution or modality-level representation divergence \cite{hsu2026anatomy, xiang2025understanding, gong2023ltumic, fathullah2024prompting}.

Building upon these broad observations, we investigate the modality gap at a fine-grained, task-specific level. By evaluating two architecturally diverse SLLMs (Qwen2.5-Omni \cite{qwen2025omni} and Phi-4-Multimodal \cite{abdin2025phi4}) across multiple reasoning categories, we reveal that the performance degradation is highly uneven. On many tasks, such as spatial navigation, syntactic ordering, and factual understanding, S2T performance is fully comparable to T2T performance. Instead, the severe modality gap is disproportionately concentrated in specific categories, most notably logical reasoning tasks that require entity tracking. For instance, on the \textit{web of lies} task, S2T accuracy across both models plummets to chance levels, while T2T baselines range up to 92\% (\Cref{tab:main_results}).

We diagnose that this localized degradation stems from the inherent nature of speech encoding. To efficiently process continuous audio sequences, SLLM encoders heavily rely on temporal pooling and downsampling mechanisms \cite{radford2023robust, defossez2022high, baevski2020wav2vec, hsu2021hubert}. While these operations effectively extract global semantic information, preserving performance on general comprehension and spatial tasks, they inevitably blur fine-grained acoustic details and discrete token boundaries. Consequently, when the model attempts to implicitly track entities and their changing properties, the smoothed-out continuous features cause it to lose track of precise details. This manifests as entity binding failures: the model understands the overall context but fails to maintain strict associations between specific entities and their changing states during implicit reasoning.

To overcome this structural bottleneck, we propose Entity-Aware Chain-of-Thought (EA-CoT), an inference-time intervention that bypasses the fragile implicit acoustic tracking. EA-CoT forces the SLLM to explicitly enumerate entities and bind them to their respective claims in the generated text space before executing step-by-step reasoning. Strikingly, enforcing this explicit structural binding allows S2T performance to recover to near-T2T levels. Furthermore, we observe that this intervention succeeds even if the SLLM's transcription of the entity name is phonetically altered or not totally accurate (e.g., mapping the spoken name ``Ka'' to ``Cass''). As long as the model establishes a consistent textual anchor in the CoT to bind properties to, the logical reasoning chain remains intact. This demonstrates that the critical failure in SLLM logical reasoning is not merely a superficial speech recognition error, but a structural semantic binding failure caused by acoustic detail loss.

Our core contributions are:
\begin{itemize}
    \item \textbf{Task-Specific Gap Diagnosis:} We demonstrate that the S2T/T2T modality gap is not a uniform cognitive deficit, but is heavily concentrated in entity-tracking tasks. Excluding these, S2T reasoning is largely comparable to T2T.
    \item \textbf{Mechanistic Explanation:} We link SLLM logical failures to the loss of acoustic details through encoder pooling, showing that blurred token boundaries disrupt implicit entity binding while global semantics remain preserved.
    \item \textbf{Effective Targeted Intervention:} We introduce EA-CoT, a lightweight inference-time intervention that forces explicit entity-property binding and serves as a causal probe of our diagnosis. Compared against structured control prompts on other tasks, it uniquely bridges the modality gap in logical reasoning (up to +24.4 pp) and remains effective even when entity names are misrecognized.
\end{itemize}

\section{Related Work}
\label{sec:related_work}

\textbf{The Speech-Text Modality Gap.} Recent end-to-end Speech Large Language Models (SLLMs) achieve strong conversational abilities but consistently underperform their text counterparts on complex reasoning benchmarks \cite{chen2024voicebench, yang2025sakura, lin2025vera, barrault2023seamlessm4t}. Prior studies have approached this modality gap primarily from a macroscopic perspective. Layer-wise and representation-level analyses attribute the performance degradation to generalized information dilution \cite{hsu2026anatomy} or modality alignment divergence \cite{xiang2025understanding}, suggesting that the continuous and redundant nature of speech tokens hinders stable late-layer decisions. Concurrent works attempt to close this gap via training-time cross-modal distillation or trajectory alignment \cite{hu2026cord, wang2026closing}. In contrast, we take a task-specific diagnostic approach, showing that the gap is not a uniform deficit but a highly localized failure concentrated in entity-centric logical reasoning.

\noindent\textbf{Entity Binding in Foundation Models.} The ``binding problem'', the cognitive capacity to robustly associate specific entities with their dynamic properties and states, is a recognized vulnerability in neural architectures \cite{bubeck2023sparks}. Mechanistic interpretability studies reveal that text-based LLMs rely on specialized internal vectors to maintain these links \cite{feng2024how}, and entity tracking remains fragile even in pure text scenarios \cite{kim2023entity, liu2024lost}. Similarly, binding failures have been shown to drive multi-object reasoning errors in Vision-Language Models \cite{campbell2024understanding}. We are the first to extend this lens to the speech modality, identifying binding failure---induced by encoder pooling that blurs the discrete token boundaries required for semantic binding---as the primary bottleneck of the SLLM modality gap.

\noindent\textbf{Chain-of-Thought for Modality Adaptation.} Chain-of-Thought (CoT) prompting fundamentally enhances LLM reasoning by externalizing intermediate deductive steps \cite{wei2022chain, zheng2023ddcot, kojima2022large, wang2022self, yao2024tree}. In the audio domain, recent multimodal CoT adaptations primarily focus on analyzing explicit acoustic events (e.g., environmental sounds) \cite{xiong2025thinking} or internalizing ASR transcripts to reduce response latency \cite{yuen2024internalizing}. However, applying generic CoT (e.g., ``Let us think step by step'') to SLLMs on logical tasks yields minimal improvement because the implicit mapping between continuous audio features and discrete entities is already corrupted. In contrast, our Entity-Aware CoT (EA-CoT) is a targeted structural intervention that projects blurred acoustic entities into a stable text space before reasoning, establishing textual anchors that survive imperfect phonetic recognition.

%% file: text/2_method.tex
\section{Method}
\label{sec:method}

\subsection{Per-Task Gap Analysis}
\label{sec:per_task_gap_analysis}
Rather than reporting a single aggregate modality gap, we evaluate speech and text inputs separately on each BBH category. We use four categories from the VoiceBench BBH split \cite{chen2024voicebench, suzgun2023challenging}: \textit{hyperbaton}, \textit{navigate}, \textit{sports understanding}, and \textit{web of lies}, totaling 1,000 items (250 per category). Each item is presented as both synthesized speech and plain text to the same model, enabling paired accuracy comparisons. This decomposition precisely reveals which reasoning abilities are most affected by the speech modality.

\begin{figure}[t]
  \centering
  \includegraphics[width=\columnwidth]{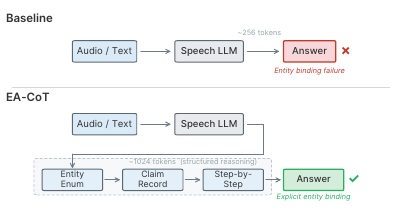}
  \caption{Method overview. The baseline generates a direct answer within 256 tokens, during which entity binding can fail for speech inputs. EA-CoT instructs the model to enumerate entities, record claims, and reason step by step within 1,024 tokens, converting implicit binding into explicit text.}
  \label{fig:method_overview}
\end{figure}

\subsection{Task-Specific Structured Chain-of-Thought}
\label{sec:task_specific_cot}
To address the entity binding bottleneck diagnosed in \Cref{sec:introduction}, we design a prompting strategy that forces the model to make the implicit binding process explicit. For \textit{web of lies}, the Entity-Aware Chain-of-Thought (EA-CoT) prompt prepends four instructions before the task input (as illustrated in \Cref{fig:method_overview}):
\begin{enumerate}
    \item \textbf{Entity enumeration.} List all people mentioned.
    \item \textbf{Claim recording.} Write down each statement linking a person to a property.
    \item \textbf{Step-by-step reasoning.} Resolve each claim in sequence.
    \item \textbf{Answer extraction.} Produce the answer in the required format.
\end{enumerate}

This forces the SLLM to externalize entity-property associations into generated text, converting fragile implicit tracking into explicit textual anchoring. EA-CoT is fully automatic: the prompt contains only these task-level instructions, and the entity list and claim record are generated by the SLLM itself during inference, without human-provided entity lists, transcripts, or oracle annotations. The maximum generation length is expanded to 1,024 tokens to accommodate the reasoning trace.
To rigorously isolate the entity binding bottleneck, we deploy structured control prompts for the remaining categories. These force explicit step-by-step reasoning without entity tracking: listing and classifying adjectives by type (\textit{hyperbaton}), sequentially tracking coordinate changes from the origin (\textit{navigate}), and identifying sports to evaluate action plausibility (\textit{sports}). These isomorphic controls ensure that EA-CoT’s disproportionate S2T gain on \textit{web of lies} reflects targeted semantic repair rather than generic CoT benefits.

\subsection{Token Budget Control}
\label{sec:token_budget_control}
Increasing the generation limit from 256 to 1,024 tokens introduces a confound, as models may improve simply by having more space to reason. To disentangle the instruction effect from the token budget effect, we evaluate a baseline with 1,024 tokens but the default instruction, denoted BL(1024). The decomposition is:
\begin{equation}
\label{eq:budget}
\Delta_{\text{Total}} = [\text{BL}_{1024} - \text{BL}_{256}]_{\text{budget}} + [\text{CoT} - \text{BL}_{1024}]_{\text{instruction}}
\end{equation}
This is applied to both speech and text inputs across both models.

%% file: text/3_experiments.tex
\section{Experiments}
\label{sec:experiments}

\subsection{Setup}
\label{sec:setup}
Two openly released, reproducible speech LLMs spanning different architectures are evaluated. Qwen2.5-Omni-7B \cite{qwen2025omni} employs a dedicated thinker module that generates internal reasoning tokens before the visible response. Phi-4-Multimodal \cite{abdin2025phi4} generates responses directly without a separate reasoning module. Both models are evaluated in their released configurations without fine-tuning.

Four categories from VoiceBench BBH \cite{chen2024voicebench} are used, each containing 250 items for a total of 1,000 samples. Each item is evaluated with both speech and text input using the same model. A format guard ensures responses match the expected answer format, with deterministic fallback extraction. Outputs without a parseable final answer are conservatively counted as incorrect. Statistical significance is assessed using McNemar's test on paired outcomes. The baseline generation budget is 256 tokens. CoT experiments use 1,024 tokens, as described in \Cref{sec:task_specific_cot}.

\begin{table*}[th]
\footnotesize
  \caption{BBH accuracy (\%) across two SLLMs (1,000 items). Scores are \textbf{S2T / T2T} to highlight the modality gap directly. BL = baseline (256 tokens). CoT = task-specific structured prompting with 1,024 tokens: EA-CoT on WOL and isomorphic structured controls on the other categories (\Cref{sec:task_specific_cot}). Best S2T result per model is in \textbf{bold}.}
  \label{tab:main_results}
  \centering
  \begin{tabular}{llccccc}
    \toprule
    \textbf{Model} & \textbf{Method} & \textbf{Overall} & \textbf{HYP} & \textbf{NAV} & \textbf{SPO} & \textbf{WOL} \\
    \midrule
    \multirow{2}{*}{\textbf{Qwen2.5-Omni}} & BL & 59.9 / 67.0 & 73.2 / 72.0 & 58.0 / 52.8 & 55.6 / 56.4 & 52.8 / 86.8 \\
     & CoT & \textbf{68.4} / 84.3 & 62.4 / 83.2 & \textbf{80.4} / 80.8 & \textbf{61.6} / 77.6 & \textbf{69.2} / 95.6 \\
    \midrule
    \multirow{2}{*}{\textbf{Phi-4-MM}} & BL & 53.6 / 66.7 & 56.4 / 61.6 & 59.2 / 58.0 & 48.0 / 55.6 & 50.8 / 91.6 \\
     & CoT & \textbf{62.7} / 77.6 & 54.8 / 77.2 & \textbf{66.4} / 82.0 & \textbf{54.4} / 64.0 & \textbf{75.2} / 87.2 \\
    \bottomrule
  \end{tabular}
\end{table*}

\begin{figure}[t]
  \centering
  \includegraphics[width=\columnwidth]{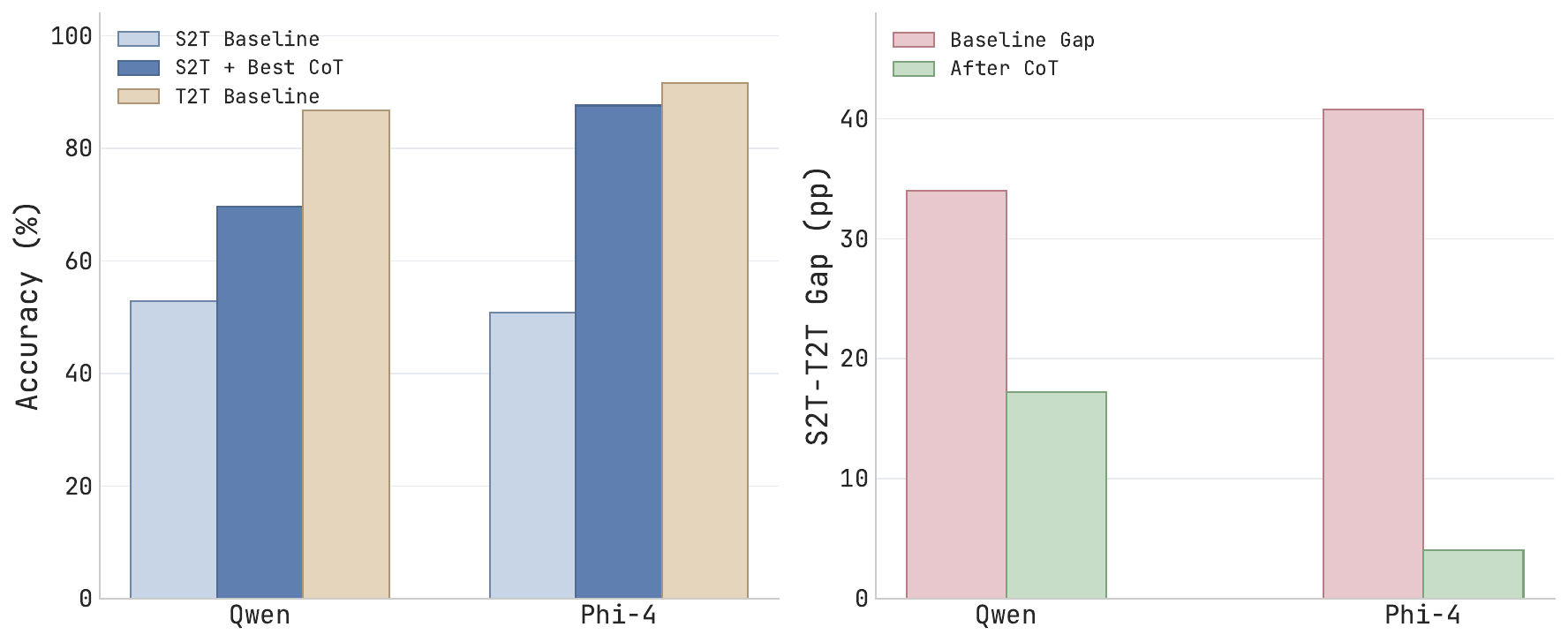}
  \caption{Speech recovery and gap reduction on \textit{web of lies}. Left: speech accuracy rises from chance toward the text baseline with EA-CoT. Right: the gap shrinks substantially across both models.}
  \label{fig:gap_reduction}
\end{figure}

\subsection{Main Results}
\label{sec:main_results}
\Cref{tab:main_results} presents the full results. As predicted by the entity binding hypothesis, \textit{web of lies} shows by far the largest baseline gap, with both models at chance level on speech despite text accuracy ranging up to 91.6\%. Most other categories show baseline gaps below 8~pp, and two categories even slightly favor speech over text. Excluding \textit{web of lies}, Qwen speech accuracy exceeds text by $+1.9$~pp, and the Phi-4 gap shrinks from 13.1 to 3.9~pp. \textit{Web of lies} alone accounts for the vast majority of the overall modality gap across both models.

Task-specific CoT improves speech-input accuracy by $+8.5$ to $+9.1$~pp overall for Qwen and Phi-4, both significant by McNemar's test. On \textit{web of lies}, entity enumeration yields $+16.4$~pp for Qwen and $+24.4$~pp for Phi-4, as shown in \Cref{fig:gap_reduction}. The magnitude of recovery scales with text-input accuracy on the same task. Phi-4 has the strongest text baseline at 91.6\% and recovers $+24.4$~pp, while Qwen at 86.8\% recovers $+16.4$~pp.

Critically, \textit{web of lies} is the only category where speech gains exceed text gains. Qwen speech improves by $+16.4$~pp versus $+8.8$~pp for text, and Phi-4 speech improves by $+24.4$~pp versus $-4.4$~pp for text. For all other categories, text benefits more from our structured controls than speech does. This asymmetry confirms that entity binding is a speech-specific bottleneck, not a general reasoning deficit.

\begin{figure}[h]
  \centering
  \includegraphics[width=0.9\columnwidth]{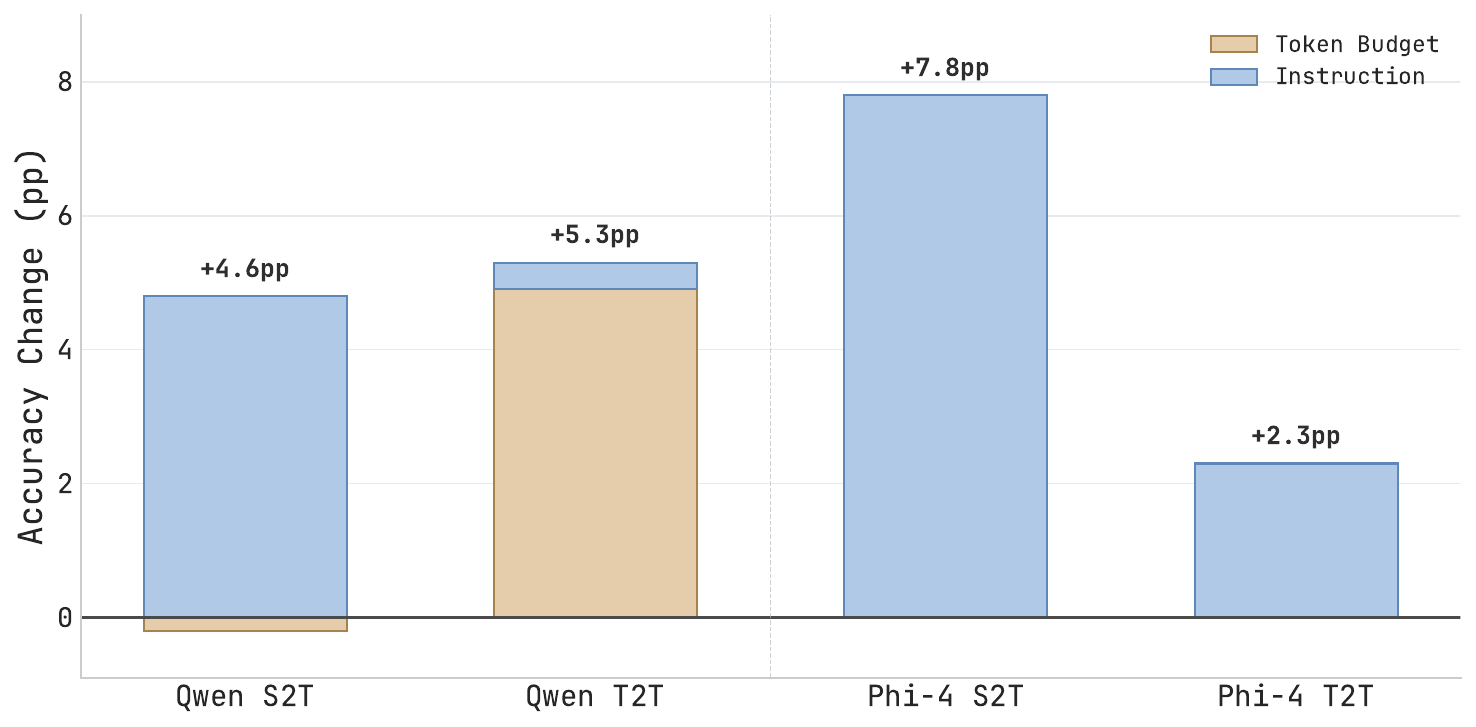}
  \caption{Decomposition of performance gains. The improvement on speech inputs is driven almost entirely by the structured EA-CoT instruction, whereas merely increasing the token budget provides negligible benefit.}
  \label{fig:token_budget}
\end{figure}

\subsection{Token Budget Control}
\label{sec:token_budget_control_exp}
Using the decomposition in \Cref{eq:budget}, we verify that the speech improvement is instruction-driven rather than a side effect of the larger token budget. Increasing the generation limit from 256 to 1,024 tokens without changing the instruction produces no improvement on speech input, with $\Delta \le 0.2$~pp for both models. For text input, the same token increase yields up to $+4.1$~pp for Qwen. Since task-specific CoT achieves $+8.5$ to $+9.1$~pp on speech, the entire gain comes from the instruction content, not from additional generation capacity. This dynamic is visually summarized in \Cref{fig:token_budget}. This asymmetry directly supports the entity binding hypothesis. Speech inputs suffer from a binding failure that extra tokens cannot resolve, but structured instructions can. Text inputs encode entities in a form the model can already bind, so more tokens enable deeper reasoning.

\begin{table}[t]
\footnotesize
  \caption{EA-CoT ablation on Qwen S2T \textit{web of lies}, 250 items, from an independent decoding run (baseline 51.6 vs.\ 52.8 in \Cref{tab:main_results}). Entity enumeration provides the largest individual contribution at 59\% of the full effect.}
  \label{tab:ablation}
  \centering
  \begin{tabular}{lccc}
    \toprule
    \textbf{Component} & \textbf{Acc (\%)} & \textbf{$\Delta$ vs BL} & \textbf{\% of full} \\
    \midrule
    Baseline & 51.6 & -- & -- \\
    + Format only & 55.6 & $+4.0$ & 23\% \\
    + Step-by-step & 59.2 & $+7.6$ & 43\% \\
    + Entity enum & 62.0 & $+10.4$ & 59\% \\
    \midrule
    Full EA-CoT & 69.2 & $+17.6$ & 100\% \\
    \bottomrule
  \end{tabular}
\end{table}

\subsection{Ablation Study}
\label{sec:ablation_study}
\Cref{tab:ablation} ablates EA-CoT components on Qwen speech-input \textit{web of lies} with 250 items. Each row tests one component independently. Step-by-step and entity enumeration both include format enforcement. Format enforcement alone provides $+4.0$~pp, indicating some baseline errors stem from formatting rather than reasoning. Adding step-by-step reasoning to the format baseline yields $+7.6$~pp total. Entity enumeration with format yields the largest gain at $+10.4$~pp, accounting for 59\% of the full EA-CoT effect and contributing $+6.4$~pp beyond format alone. The full combination achieves $+17.6$~pp, confirming that explicit entity listing is the most critical component, consistent with the entity binding hypothesis.

%% file: text/4_analysis.tex
\section{Analysis}
\label{sec:analysis}

\textbf{Semantic Binding vs. Acoustic Recognition.} If the speech-text gap simply arose from mishearing names (ASR errors), corrupting names in text inputs should trigger a similar performance collapse. However, replacing 100\% of person names with random strings in T2T \textit{web of lies} (\Cref{tab:corruption}) reduces Qwen's accuracy by only 3.6~pp, accounting for barely 11\% of the 34~pp S2T gap. This confirms that the bottleneck lies in maintaining semantic associations across reasoning steps, not in acoustic perception. Further corroborating this, \Cref{tab:example} illustrates that EA-CoT successfully tracks the logical chain even when the model transcribes spoken names inconsistently (e.g., mapping ``Ka'' to ``Cass'' and ``Inga'' to ``Ignatia''). By externalizing entities into explicit text, EA-CoT establishes stable anchors that preserve reasoning regardless of phonetic precision.

\begin{table}[th]
\footnotesize
  \caption{Entity corruption on T2T \textit{web of lies}, 250 items, Qwen. Even 100\% name corruption reduces accuracy by only 3.6~pp, far less than the 34~pp S2T gap.}
  \label{tab:corruption}
  \centering
  \begin{tabular}{cccc}
    \toprule
    \textbf{Corruption} & \textbf{BL (256)} & \textbf{EA-CoT} & \textbf{$\Delta$} \\
    \midrule
    0\% & 89.6 & 95.6 & $+6.0$ \\
    25\% & 89.2 & 94.4 & $+5.2$ \\
    50\% & 84.8 & 93.2 & $+8.4$ \\
    75\% & 88.4 & 94.0 & $+5.6$ \\
    100\% & 86.0 & 92.0 & $+6.0$ \\
    \bottomrule
  \end{tabular}
\end{table}

\begin{table}[t]
\footnotesize
  \caption{Example S2T output on \textit{web of lies}, Qwen. EA-CoT enumerates entities and tracks the chain correctly despite name transcription differences.}
  \label{tab:example}
  \centering
  \begin{tabular}{p{0.95\columnwidth}}
    \toprule
    \textbf{Input (speech):} Alexis tells the truth. Shenna says Alexis tells the truth. Yolanda says Shenna tells the truth. Ka says Yolanda tells the truth. Inga says Ka tells the truth. Does Inga tell the truth? \textbf{Gold: Yes} \\
    \midrule
    \textbf{BL (256 tok):} No. \hfill $\boldsymbol{\times}$ \\
    \midrule
    \textbf{EA-CoT (1024 tok):} The people mentioned are Alexis, Shanna, Yolanda, Cass, and Ignatia. Alexis tells the truth. Shanna says Alexis tells the truth, so Shanna is telling the truth. Yolanda says Shanna tells the truth, so Yolanda is telling the truth. Cass says Yolanda tells the truth, so Cass is telling the truth. Ignatia says Cass tells the truth, so Ignatia is telling the truth. The answer is yes. \hfill $\boldsymbol{\checkmark}$ \\
    \bottomrule
  \end{tabular}
\end{table}

\textbf{Intervention Specificity.} EA-CoT specifically targets semantic binding. Applying a generic CoT prompt (``Let us think step by step'') to Qwen yields only a marginal $+2.4$~pp gain on \textit{web of lies}, proving that explicit entity enumeration ($+16.4$~pp) is the critical catalyst. Conversely, when EA-CoT is applied to MMSU \cite{wang2025mmsu}, an acoustic-heavy benchmark where critical paralinguistic cues are inherently lost in text, it yields no improvement (\Cref{tab:mmsu}). Because MMSU answers hinge on acoustic evidence rather than entity-property bookkeeping, the absence of MMSU gains supports the specificity of our diagnosis rather than weakening it. This validates that our intervention strictly repairs semantic structural binding rather than acting as a generic reasoning enhancer.

\begin{table}[th]
\footnotesize
  \caption{Specificity validation: BBH (semantic) vs. MMSU (acoustic) \cite{wang2025mmsu}. On MMSU, where tasks depend on acoustic cues, the gap reverses. Entity enumeration produces no gain on MMSU, confirming EA-CoT targets semantic binding specifically.}
  \label{tab:mmsu}
  \centering
  % \resizebox{\columnwidth}{!}{%
  \begin{tabular}{lcc}
    \toprule
    \multirow{2}{*}{\textbf{Model}} & \textbf{BBH (TTS, 1K)} & \textbf{MMSU (natural, 2.4K)} \\
    & \textbf{S2T / T2T ($\Delta$)} & \textbf{S2T / T2T ($\Delta$)} \\
    \midrule
    \textbf{Qwen} & 59.9 / 67.0 ($-7.1$) & 80.1 / 49.0 ($+31.0$) \\
    \textbf{Phi-4} & 53.6 / 66.7 ($-13.1$) & 71.9 / 47.5 ($+24.5$) \\
    \bottomrule
  \end{tabular}%
  % }
\end{table}

\textbf{Recovery Requires Text-Level Capability.} EA-CoT's efficacy inherently depends on foundational text-level capability. On \textit{web of lies}, the degree of S2T recovery scales monotonically with T2T accuracy. Both models boast strong T2T baselines ($>86\%$) and recover substantially in speech ($+24.4$ and $+16.4$~pp). Output-level analysis reveals the mechanism: Phi-4's speech baseline mostly produces bare answers (97\% of outputs under 30 characters) despite generating full step-by-step chains on text input. CoT prompting successfully restores this reasoning activation. By bypassing modality-induced bottlenecks, CoT can effectively externalize the binding capability that the SLLM already possesses in its text representation.

\textbf{Inference Latency Trade-off.} While EA-CoT effectively repairs the modality gap, it inherently increases inference latency. Externalizing the binding process significantly expands the generated token sequence, explicitly trading generation speed for logical accuracy. For real-time spoken dialogue systems, this overhead is a practical limitation. Future optimizations could explore distilling these explicit reasoning traces back into the model's implicit representations to maintain structural robustness without the autoregressive delay.

%% file: text/5_conclusion.tex
\section{Conclusion}
\label{sec:conclusion}

Entity binding failure is the primary driver of the S2T/T2T modality gap in complex reasoning. We demonstrate that SLLMs maintain robust performance on spatial and factual tasks, but tasks requiring continuous entity tracking collapse to chance-level accuracy on speech inputs. To causally validate this diagnosis, we introduce Entity-Aware Chain-of-Thought (EA-CoT), a lightweight inference-time intervention that bypasses fragile implicit acoustic tracking by forcing models to explicitly enumerate and bind entities in text before reasoning. EA-CoT bridges the S2T gap by up to $+24.4$~pp on entity-centric tasks, even when entities are phonetically misrecognized, evidence that the reasoning capability exists, but speech input fails to elicit it reliably.

While highly effective, EA-CoT is an inference-time prompting strategy that roughly triples token generation and latency. Additionally, our evaluation relies on TTS-generated speech and 7B-class models; real-world acoustic noise and larger architectures may introduce varied compound effects. Future work should explore representation-level alignments, such as synchronizing cross-modal entity binding subspaces \cite{feng2024how}, to stabilize continuous entity tracking without the overhead of explicit prompting.